\documentclass[runningheads]{llncs}

\usepackage{amssymb}
\usepackage{booktabs}
\usepackage{todonotes}
\usepackage{amsmath}
\usepackage[all]{nowidow}
\usepackage{needspace}
\usepackage{hyperref}
\usepackage{graphicx} 
\usepackage{import}
\usepackage{caption}
\usepackage{float}
\usepackage{color}
\usepackage{transparent}
\usepackage{subcaption}

\usepackage{tikz}
\usetikzlibrary{arrows.meta, positioning, patterns} 

\begin{document}




\title{Towards Hierarchical Structure Understanding of Newspaper Images} 
\titlerunning{Understanding the Structure of Newspaper Images}

\author{William Mocaër \inst{1} \and Solène Tarride\inst{3} \and Thomas Constum\inst{1} \and Merveilles Agbeti-Messan\inst{1} \and Tom Simon\inst{1} \and Clément Chatelain\inst{2} \and Stéphane Nicolas\inst{1} \and Pierrick Tranouez\inst{1} \and Sébastien Cretin\inst{4} \and Thierry Paquet\inst{1}} 

\institute{LITIS UR4108, University of Rouen Normandy, France \\
\email{\{thomas.constum, komlan-epe-nsin.agbeti-messan, tom.simon, stephane.nicolas, pierrick.tranouez, thierry.paquet\}@univ-rouen.fr}
\and LITIS UR4108, INSA of Rouen Normandy, France \\
\email{clement.chatelain@insa-rouen.fr}
\and Teklia, Paris, France\\
\email{starride@teklia.com}
\and Bibliothèque nationale de France, Paris, France \\
\email{sebastien.cretin@bnf.fr}
}

\authorrunning{W. Mocaër et al.}
\maketitle

\begin{abstract}
Understanding newspaper images remains a challenging task due to their complex, nested hierarchical structures and dense, heterogeneous layouts. In this paper, we explore two complementary approaches for newspaper structure understanding.\\
First, we present a modular bottom-up pipeline that combines state-of-the-art open-source models: YOLO for layout detection, LayoutReader for reading order prediction, and a custom algorithm for article segmentation. This approach leverages existing robust components while maintaining flexibility and interpretability.\\
Second, we introduce Tiramisu (Tiered Transformers for Hierarchical Structure Understanding), a novel end-to-end transformer-based architecture that explicitly models document hierarchy through an iterative tiered process. Tiramisu performs section and article separation, block localization, semantic categorization, and reading order prediction using highly parallelized attention mechanisms.\\
Finally, we release \textbf{Finlam La Liberté}, a new dataset designed specifically for evaluating hierarchical information retrieval in historical newspapers. Experimental results demonstrate the effectiveness of both approaches in reconstructing complex newspaper hierarchies, with comparative analysis highlighting their respective strengths for scalable document digitization. 
The Tiramisu training code, including the synthetic newspaper generator, is available at
\url{https://git.litislab.fr/tiramisu/tiramisu-newspaper-articles-extractor}.

\end{abstract}




\section{Introduction}


The large-scale digitization of historical newspapers has opened up new opportunities for digital humanities, media analysis, and information retrieval. However, the structural complexity and sheer volume of newspaper archives pose significant challenges for automated document understanding. Newspaper pages typically feature high-resolution images, diverse layouts, and a dense organization of heterogeneous content, such as sections, articles, titles, advertisements, and illustrations, often embedded within a nested hierarchical structure.


Existing approaches typically fall into two categories: (i) task-specific methods that address isolated problems (e.g., semantic labeling~\cite{SOULLARD2020}, article separation and reading order~\cite{Girdhar2023,Sun2024}); and (ii) complex pipeline systems~\cite{Palfray2012,Tranouez2015,Zeni2017,Kettunen2019,Kontonasios2025,hip23rezanezhad} involving OCR/OLR, physical layout analysis, and logical reconstruction, often relying on handcrafted rules. While effective in constrained scenarios, these methods struggle with generalization, scalability, and joint modeling of layout and semantics.

Recent advances in vision-language models have shown promise for document understanding. Transformer-based architectures such as Donut~\cite{donut2022}, Nougat~\cite{nougat2023}, and Éclair~\cite{karmanov2025eclair} have demonstrated strong performance on tasks like document parsing, structured prediction, and layout-based reasoning. However, these models have not yet been applied to the specific challenges of newspaper analysis, which requires modeling deep hierarchical structures, long-range dependencies, and spatial reasoning over large and densely packed pages.


In this paper, we investigate two distinct strategies for newspaper image analysis and recognition. 
The bottom-up strategy assembles a pipeline of specialized models: YOLO \cite{yolo} detects and categorizes blocks, which are then organized into coherent reading sequences using LayoutReader \cite{wang2021-layoutreader}. The top-down strategy employs a transformer-based architecture that directly learns to identify hierarchical organization, proceeding from high-level sections down to fine-grained elements.

Our contributions are as follows:
\begin{itemize}
    \item We propose an end-to-end transformer-based model tailored for a top-down structured newspaper analysis;
    \item We explore a bottom-up approach for newspaper structure extraction and reading order detection; 
    \item We propose a comprehensive evaluation framework for structured information extraction on historical newspapers;
    \item We release a newspaper dataset; 
    \item We provide a comprehensive comparison of both approaches with their respective strengths and weaknesses. 
\end{itemize} 

This paper is structured as follows. Section~\ref{sec:related_work} examines related work, and Section~\ref{sec:newspapers} introduces the Newspaper Structure Representation. Section~\ref{sec:pipeline} presents our bottom-up approach and Section~\ref{sec:tiramisu} presents the top-down Tiramisu architecture and its multi-pass strategy. We introduce the Finlam La Liberté dataset in Section ~\ref{sec:dataset}, and the evaluation results in Section~\ref{sec:evaluation}. Finally, Section~\ref{sec:conclusion} concludes the paper and discusses future work.

\begin{figure*}[htbp]
  \centering
  \includegraphics[width=\linewidth]{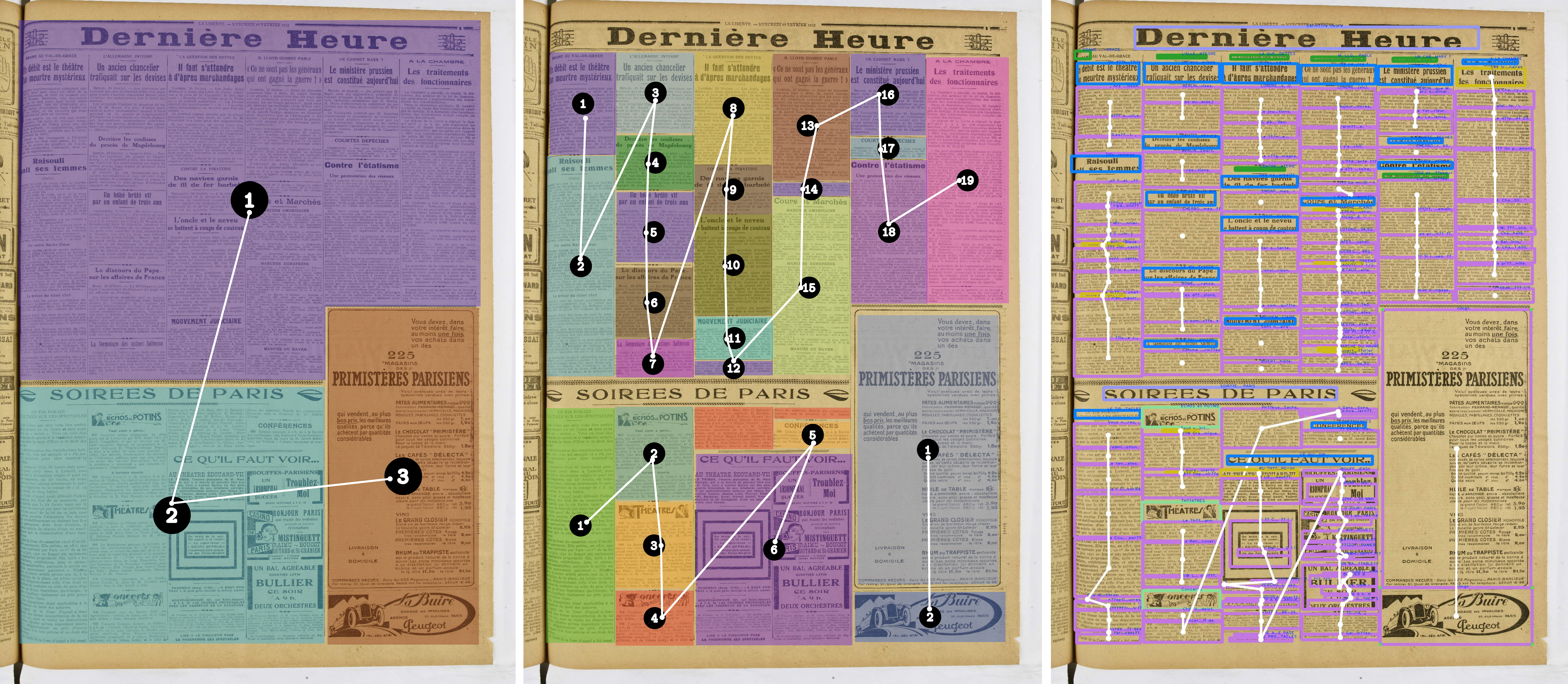}
  \caption{Hierarchical representation of newspapers. From left to right: (a) Section level - identification of sections along with their reading order; (b) Article level – segmentation and classification of articles with reading order and section membership preserved; (c) Block level - detection and classification of individual content blocks (e.g., paragraph, image, title, table), along with their local reading order within each article.}
  \label{fig:task_tiramisu}
\end{figure*}

\section{Related work}
\label{sec:related_work}

Until now, historical newspaper analysis has been investigated mostly using task-specific approaches or pipeline-based systems. Some approaches focus on isolated tasks, such as semantic segmentation from images, often using semi-automatically annotated data. Pipeline methods offer more comprehensive outputs by combining OCR/OLR with physical layout analysis (e.g., lines, columns, separators), but depend heavily on handcrafted rules. In parallel, large vision-language models like Donut~\cite{donut2022} show promising results on general document tasks, though they remain unexplored in the context of historical newspapers.

\subsection{Task-specific approach for Newspaper Analysis}
Specialized models have been developed to address individual sub-tasks in newspaper analysis, each targeting specific aspects of document structure.

Soullard et al.\cite{SOULLARD2020} introduce a fully convolutional neural network (FCN) designed for \textbf{pixel-level semantic labeling} of historical newspaper images. The primary challenge addressed is the detection of various types of block, such as advertisements, images, titles, and paragraphs. To tackle this, the authors focus on a multi-scale analysis strategy, enabling the model to capture and select relevant features across different resolutions for improved semantic understanding. 

LayoutReader~\cite{wang2021-layoutreader} addresses the \textbf{reading order} prediction task in visually complex documents as a sequence-to-sequence problem, leveraging a transformer architecture that integrates textual content with spatial and structural layout features. It is designed to process heterogeneous document layouts, including forms, receipts, and multi-column layout. LayoutReader assumes pre-detected content and focuses solely on determining structural organization.

STRAS~\cite{Girdhar2023} and LIAS~\cite{Sun2024} address article separation and intra-article \textbf{reading order} in historical newspapers using different strategies. Built on top of an OCR, STRAS relies on textual cues by computing contextual embeddings of text regions and grouping them based on similarity and predefined rules. In contrast, LIAS is layout driven: it detects separator lines, reconstructs the document information flow, and segments articles through a neural classifier operating on layout features. While STRAS emphasizes semantic similarity between text blocks, LIAS focuses on geometry, the structural layout and the sequential organization to infer article boundaries.

Several other approaches widely used in the digital humanities community rely on pipeline-based systems for the analysis of newspapers~\cite{hip23rezanezhad,Palfray2012,Tranouez2015,Zeni2017,Kettunen2019,Kontonasios2025}. These methods typically start from OCR or OLR outputs, often annotated semi-automatically, and proceed through a sequence of steps involving physical layout analysis, such as the detection of lines, columns, and visual separators, followed by the construction of a logical structure (e.g., articles or reading order) from these elements. A significant portion of these systems depends on handcrafted rules to guide segmentation and classification decisions.



\subsection{Large Vision Model for Document image understanding.}

Recent large vision models offer a unified alternative to task-specific pipelines, enabling \textbf{end-to-end document processing}.

Donut~\cite{donut2022} pioneered OCR-free document understanding using a Vision Encoder-Decoder architecture (Swin Transformer encoder, autoregressive decoder) trained on diverse document types. It directly outputs structured text without external OCR engines. Nougat~\cite{nougat2023} applies a similar architecture to scientific PDFs, transcribing them into Markdown while preserving structure. Éclair~\cite{karmanov2025eclair} targets more challenging structured extraction tasks and adds spatial localization through learnable coordinate tokens predicted alongside text.




More recently, \textbf{Large Vision Language Models} (LVLM) such as Qwen \cite{qwen} and Gemma \cite{gemma} have emerged as relevant general-purpose systems able to combine textual and image modalities to process document images in place of the traditional OCR pipelines. While large vision-language models have shown strong performance on receipts, forms, or scientific papers, applying them directly to newspaper pages remains highly challenging. These documents typically combine high-resolution images with very dense layouts, overlapping or nested elements, and a complex logical structure that requires hierarchical modeling of sections and articles. 
Furthermore, real-world applications require models to predict not only the document structure but also the spatial localization of elements, enabling end users to precisely locate retrieved information and allowing curators and digitization teams to perform quality control on digitization outputs.

\subsection{Commercial services for Newspaper Recognition}

Arcanum \cite{arcanum_segmentation} provides a commercial newspaper digitization service that performs layout analysis, article segmentation, OCR, and metadata extraction for historical newspaper collections. This service was used to process over 100 million pages across multiple European archives.
While the underlying pipeline is proprietary and operates as a black box, we use it as a baseline for evaluating our proposed approaches.


 \section{Newspaper Structure Representation}
\label{sec:newspapers}

In heritage institutions, digitized newspapers are commonly stored in
complementary METS and ALTO formats: the METS file encodes the logical
structure of the issue (grouping into articles, sections, and semantic zone
typing), while one ALTO file per page provides the physical layout
(bounding boxes of zones and lines) and the OCR transcription. In this
work, we jointly exploit both formats and convert them into a unified JSON
representation used by all our models. METS consistently separates each
newspaper into two main components: the Title Section (TS) and the Content
Section (CS). The Title Section refers to the top banner of the first page, which contains general information such as the name of the newspaper, the date, the editor, etc. The Content Section includes the remainder of the newspaper, structured into sections, articles, and content blocks (text, images, captions, etc.), allowing for a detailed and coherent representation of the newspaper's logical layout and physical information such as bounding boxes. An example of the hierarchical structure of a "Content Section" page is shown in Figure~\ref{fig:task_tiramisu}.


In this work, we use a simplified representation of a newspaper, retaining only the information relevant to our extraction objectives. This representation preserves the essential structural and spatial elements, such as sections, articles, blocks, their hierarchical relationships and reading order, and their positions on the page, while discarding unnecessary layout details. We describe this representation below.



Formally, we define a newspaper structure as a couple \(\mathcal{N} = (\texttt{TS}, \texttt{CS}) \) 
where \( \texttt{TS} = [b^{(\texttt{TS})}_1, b^{(\texttt{TS})}_2, \dots, b^{(\texttt{TS})}_{N_{\texttt{TS}}}] \) is the \emph{Title Section} made of a sequence of blocks, and  \( \texttt{CS} = [ S_1, S_2, \dots, S_{N_S} ] \) is the \emph{Content Section} made of a sequence of sections.

Each section \( S_i \) is a sequence of articles  \(S_i = [ A_{i1}, A_{i2}, \dots, A_{iN_A^{(i)}}] \), 
and each article \( A_{ij} \) is a tuple consisting of a class label and two sequences of blocks 
\(A_{ij} = \left( \texttt{class}^{\text{art}}_{ij}, \texttt{HEAD}_{ij}, \texttt{BODY}_{ij} \right)\) with the following properties:

\begin{itemize}
    \item \( \texttt{class}^{\text{art}}_{ij} \in \mathcal{C}_{\texttt{art}},\quad\text{with} \quad\mathcal{C}_{\texttt{art}}= \{\text{article}, \text{ad}, \text{freead}\} \)
    \item \( \texttt{HEAD}_{ij} = [b^{(\texttt{head})}_{ij1}, b^{(\texttt{head})}_{ij2}, \dots, b^{(\texttt{head})}_{ijN_H}] \)
    \item \( \texttt{BODY}_{ij} = [b^{(\texttt{body})}_{ij1}, b^{(\texttt{body})}_{ij2}, \dots, b^{(\texttt{body})}_{ijN_B}] \)

\end{itemize}

Each block \( b \) is defined as a tuple \( b_{i,j,k}^{(\cdot)} = \left( \mathbf{t}, \texttt{class}^{\text{block}}_{i,j,k}, \mathbf{p} \right)\) with the following properties:


\begin{itemize}
    \item \( \mathbf{t} \in \mathcal{V}^T \): a sequence of \( T \) tokens from the vocabulary \( \mathcal{V} \)
    \item \( \texttt{class}^{\text{block}} \in \mathcal{C}_{(\cdot)} \), where \( (\cdot) \in \{\texttt{head}, \texttt{body}, \texttt{TS}\} \) indicates the context in which the block appears:
    \begin{itemize}
        \item \( \mathcal{C}_{\texttt{head}} = \{\text{section-title},\ \text{section-subtitle},\ \text{section-illustrated-text},\ \text{title},\\ \text{subtitle},\ \text{illustratedtext}\} \)
        \item \( \mathcal{C}_{\texttt{body}} = \{\text{text},\ \text{illustration}, \text{caption}, \text{table},\  \text{author}, \text{insideheading}\} \)
        \item \( \mathcal{C}_{\texttt{TS}} = \{\text{header-title}, \text{header-text},\ \text{illustration}\}\)
    \end{itemize}
    \item \( \mathbf{p} = (x, y, w, h) \in \mathbb{R}^4 \): the spatial coordinates of the block
\end{itemize}

Figure~\ref{fig:task_tiramisu} provides a visual instantiation of this representation on a real page.

Following this representation, we aim to extract the structure of each newspaper page according to this hierarchical representation.

\section{Bottom-up pipeline}
\label{sec:pipeline}
We have assembled a baseline system dedicated to scanned-newspaper recognition by combining a set of open-source state-of-the-art models. This baseline is structured as a modular pipeline that integrates specialized components for each stage of the task.  

\begin{figure}
    \centering
    \includegraphics[width=0.75\linewidth]{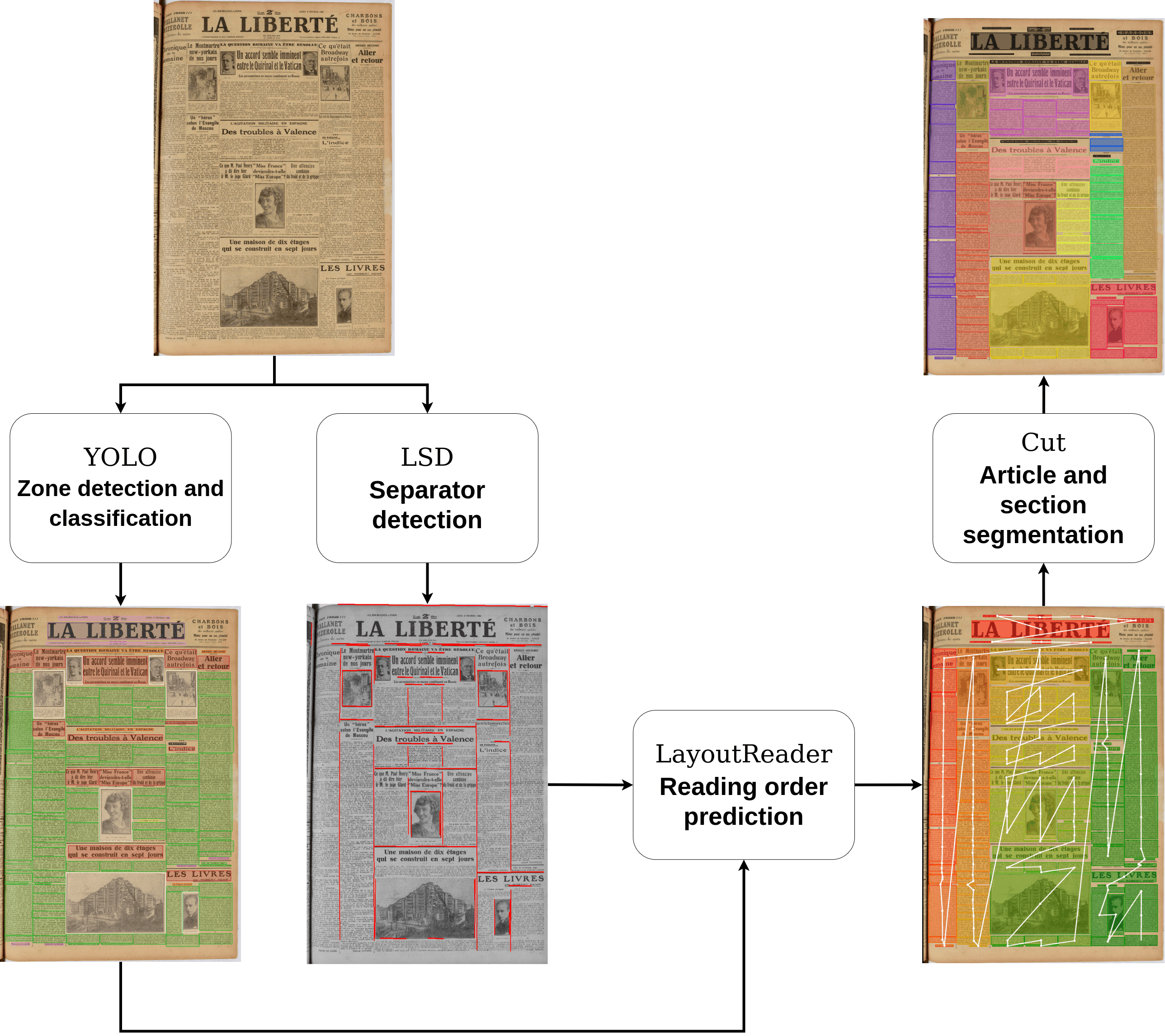}
    \caption{Illustration of the sequential pipeline. First, layout analysis is performed with YOLO to detect blocks and LSD to identify vertical and horizontal separators. The blocks are then sorted in reading order by LayoutReader. Finally, article separation is done by a custom cut algorithm.}
    \label{fig:pipeline}
\end{figure}

\subsection{Block detection and classification}

The first step of the proposed pipeline consists in detecting and classifying all the unit blocks in a given page. 
For this task, we have selected the YOLO26 model for its performance and speed \cite{yolo26}. 

The model was trained on images resized to 1024 pixels with a batch size of 8. It is trained for 250 epochs using early stopping and a patience of 100 epochs. The best performance was obtained at epoch 116, after 15 hours of training. Data augmentation is used to improve generalization. Zones of type ADVERTISEMENT and FREEAD were ignored during training, so that YOLO can focus on article content. 

During inference, zones with low confidence ($< 0.2$) are discarded, and overlapping blocks are removed using the Non-Maximal Suppression (NMS) algorithm.

\subsection{Reading order}

To restore the reading order, the pipeline employs LayoutReader, which takes an unordered set of detected blocks (predicted by YOLO) and outputs their predicted reading sequence. 

The model receives the following input features: 
\begin{itemize}
    \item block bounding box coordinates, normalized between 0 and 1000;
    \item the predicted class, represented by an integer.
\end{itemize}

Additionally, we incorporate separator information detected using OpenCV's Line Segment Detection (LSD)\cite{lsd} method to assist in identifying column and section boundaries, as our preliminary experiments showed it improves reading order quality.

During training, input zones are shuffled, and LayoutReader learns to order them. The model was fine-tuned starting from the \texttt{hantian/layoutreader} checkpoint in HuggingFace\footnote{\url{https://huggingface.co/hantian/layoutreader}}. It was trained for 30 epochs, with a batch size of 8 and a learning rate of 5e-5.

During inference, the zones are pre-sorted following a left-to-right and top-to-bottom reading order before applying LayoutReader.

\subsection{Article and section segmentation}

Finally, based on the representation adopted in section 3, a custom hierarchy updater organizes the output by ordering zones according to their predicted reading sequence and segmenting articles by identifying article-title regions.

The algorithm iterates through regions in reading order and creates a new article (or section) whenever an ARTICLE-TITLE (or SECTION-TITLE) zone or group of zones appears.






\section{A top-down model for hierarchical information extraction}
\label{sec:tiramisu}

\subsection{Architecture}

\begin{figure*}[h]
  \centering
  \includegraphics[width=\textwidth]{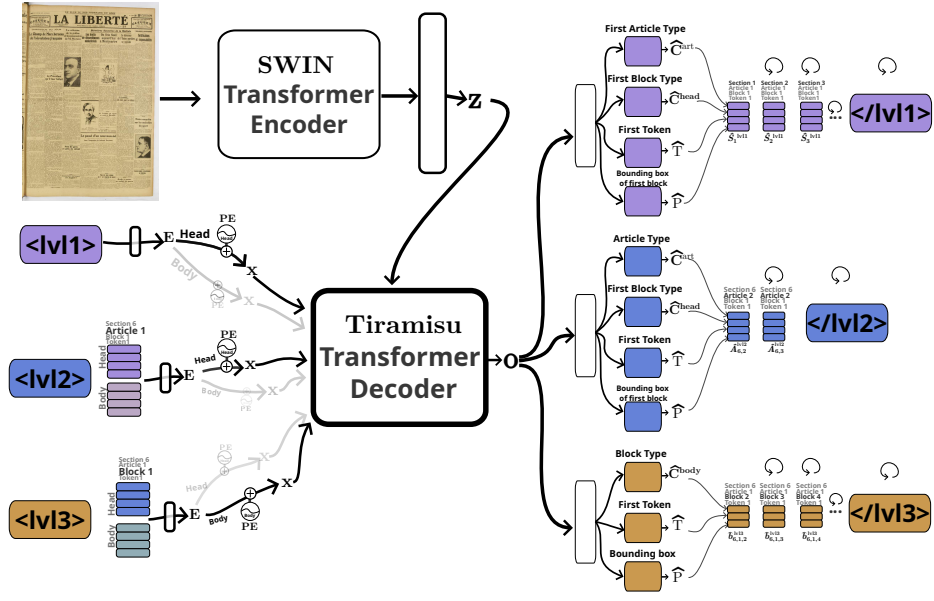}
  \caption{Overview of the TIRAMISU architecture, and its multi-pass decoding scheme guided by level-specific prompts (e.g., \texttt{<lvl1>}, \texttt{<lvl2>}, \texttt{<lvl3>}).}
  \label{fig:archi}
\end{figure*}

Inspired by Donut~\cite{donut2022} and Nougat \cite{nougat2023}, we have designed a specific architecture that extracts each information of the hierarchical representation conditioned on the previous representation level. The architecture is made of Tiered Transformers for Hierarchical Structure Understanding (TIRAMISU). It consists of a Swin transformer~\cite{swin2021} as the encoder that extracts features from the image and a transformer decoder. The Swin encoder takes as input the image $i \in \mathbb{R}^{H\times W\times 3}$ and produces a set of features of size $z \in \mathbb{R}^{D\times d}$, where $D$ is proportional to $H$ and $W$ and will be used as image tokens by the decoder, and $d$ is the embedding size. The transformer decoder takes as input the feature map $z$ from the encoder, and the sequence of structured
elements obtained at the previous level of the hierarchy (e.g., \texttt{<lvl1>}, \texttt{<lvl2>}, \texttt{<lvl3>}, detailed in \ref{multipass-process}). After the core Transformer blocks, the decoder branches into multiple specialized heads: article classification, block classification, block coordinate regression, and token prediction. Each head is responsible for a specific type of output and is used depending on the stage of the decoding process. Figure~\ref{fig:archi} gives on overview of the TIRAMISU architecture with its hierarchies of decoder heads. Using the multi-pass decoding process, the system produces the desired structure of the newspapers $\mathcal{N}$.

\subsection{Multi-pass decoding process}
\label{multipass-process}
TIRAMISU aims to extract the entire structure $\mathcal{N}$ from the image. This extraction is based on an original multi-pass decoding process following the hierarchical top-down approach, where each pass is responsible for extracting a specific structural level of the document, from sections to tokens, along with its associated attributes such as class labels, bounding boxes, and token content. 




The decoding process follows four different hierarchical pass. Pass~1 extracts sections with the first block and the first token of each section; Pass~2 identifies articles within each section, again focusing on their first blocks and first tokens; Pass~3 retrieves all blocks within each article, with their first tokens; and Pass~4 extracts the token content of each block. 

At each decoding pass, the input consists of a sequence of \( \texttt{seq} \) structured elements, each associated with multiple attributes: tokens, article-level class, block-level classes (for \texttt{HEAD} and \texttt{BODY}), and bounding box coordinates. The raw inputs are represented as follows, with \((\cdot) \in \{\texttt{head},\ \texttt{body}\}\).:

\begin{itemize}
  \item \(\mathbf{T}^{(\cdot)} \in \mathbb{N}^{\texttt{seq} \times 1}\): token IDs,
  \item \(\mathbf{C}^{\texttt{art}} \in \mathbb{N}^{\texttt{seq} \times 1}\): article-level class IDs in \([1, |\mathcal{C}_{\texttt{art}}|]\),
  \item \(\mathbf{C}^{(\cdot)} \in \mathbb{N}^{\texttt{seq} \times 1}\): block-level class IDs in \([1, |\mathcal{C}_{(\cdot)}|]\),
  \item \(\mathbf{P}^{(\cdot)} \in \mathbb{R}^{\texttt{seq} \times 4}\): normalized bounding box coordinates.
\end{itemize}


Each of these inputs is projected into a latent embedding space:
\begin{equation}
\begin{aligned}
  \mathbf{E}^{\mathbf{T}^{(\cdot)}} &\in \mathbb{R}^{\texttt{seq} \times d^{\text{token}}}, \\
  \mathbf{E}^{\mathbf{C}^{\texttt{art}}} &\in \mathbb{R}^{\texttt{seq} \times d^{\text{class}_{\texttt{art}}}}, \\
  \mathbf{E}^{\mathbf{C}^{(\cdot)}} &\in \mathbb{R}^{\texttt{seq} \times d^{\text{class}_{\texttt{block}}}}, \\
  \mathbf{E}^{\mathbf{P}^{(\cdot)}} &\in \mathbb{R}^{\texttt{seq} \times d^{\text{coords}}}.
\end{aligned}
\end{equation}

Note that tokens and coordinates share the same embedding weights for both \texttt{head} and \texttt{body}.

The resulting embeddings are concatenated and projected to a unified dimension \( d \) via a learned linear transformation:
\begin{equation}
\mathbf{E} = \mathcal{W}^{\text{proj}} \cdot \texttt{Concat}(\mathbf{E}^{\mathbf{T}^{\texttt{head}}},\ 
\mathbf{E}^{\mathbf{T}^{\texttt{body}}},\ 
\mathbf{E}^{\mathbf{C}^{\texttt{art}}},\ \mathbf{E}^{\mathbf{C}^{\texttt{head}}},\ \mathbf{E}^{\mathbf{C}^{\texttt{body}}},\ \mathbf{E}^{\mathbf{P}^{\texttt{head}}},\  \mathbf{E}^{\mathbf{P}^{\texttt{body}}}) + \mathbf{b}^{\text{proj}}
\end{equation}
where \( \mathbf{E} \in \mathbb{R}^{\texttt{seq} \times d} \) is the final embedding representation.

For training, we employ the cross-entropy loss for all classification tasks, including token prediction \( \hat{\mathbf{T}} \), article class \( \hat{\mathbf{C}}^{\text{art}} \), and block classes (\( \hat{\mathbf{C}}^{\text{head}} \), \( \hat{\mathbf{C}}^{\text{body}} \)). For the prediction of spatial coordinates \( \hat{\mathbf{P}} \), we use a combination of the Generalized Intersection over Union (GIoU) loss and Mean Squared Error (MSE): $\mathcal{L}_{\mathrm{bbox}}=\alpha \cdot \mathrm{GIoU}+(1-\alpha) \cdot \mathrm{MSE} \cdot \psi_{\mathrm{MSE}}$

\Needspace{6\baselineskip}
This multipass system offers several key advantages:
\begin{enumerate}
    \item Attribute prediction at specific levels: this approach enables the prediction of attributes at certain levels, such as article classes at level 2 (article, ad, or free-ad), and block classes and localization at level 3.
    \item Local context: Given a specific level, the context only considers other entities at the same level (previous sections, previous articles, etc.), resulting in a highly relevant and relatively short context.
    \item Parallelization capability: From level 2, the process can be easily parallelized. For example, at level 2, when extracting articles within sections, decoding can be performed in parallel for all sections. At level 4, token extraction can be parallelized across all blocks.
\end{enumerate}

\subsection{Pass-specific prompting}

To enable multi-pass processing, we design a prompting strategy that provides the decoder with distinct contextual information for each pass. Specifically, we introduce special prompt tokens \texttt{<lvl1>}, \texttt{<lvl2>}, \texttt{<lvl3>}, and \texttt{<lvl4>}, each corresponding to a decoding pass. These tokens are added to the vocabulary and prepended to the input of the decoder to signal the current pass. The model is also trained to generate their corresponding closing tokens (e.g., \texttt{</lvl1>}) to mark the end of a given pass.

A key component of this approach—contributing to the hierarchical reasoning ability of the model—is the use of contextual prompts derived from the outputs of previous passes. For example, in pass 2 which aims to extract articles within a section, each decoding instance uses a \texttt{<lvl2>} prompt followed by one of the section blocks predicted in pass 1. If pass 1 produces $N_S$ sections, then pass 2 will be applied $N_S$ times, once for each predicted section.

Formally, let $\hat{S}_i^{\texttt{lvl1}}$ denote the $i$-th section predicted in pass 1, represented by its first block. Then the prompt for pass 2, decoding the section $i$, is: \( \texttt{<lvl2>},\ \hat{S}_i^{\texttt{lvl1}}\)


The decoder receives this prompt along with the encoder output $\mathbf{z}$, allowing it to generate all articles within $\hat{S}_i^{\texttt{lvl1}}$. This process is similarly applied in subsequent passes, using the outputs of pass $k-1$ to build the prompts for pass $k$. This mechanism naturally supports a hierarchical, top-down decoding structure.



\subsection{Implementation and training}

In this work we experimented with the following architecture: $d=1024$,$d^{token}= 1024$ ,$d^{class_{art}}=8$,$d^{class_{block}}=16$, $d^{coords}=8$. For simplicity, blocks of class \emph{Advertisement} and \emph{Free ad} are excluded from the training and evaluation experiments reported in this paper. To address the limitations of our annotated dataset, which contains annotations of moderate quality and limited variability, we developed a synthetic data generator tailored to newspaper documents. This tool serves two main purposes: (i) it compensates for incomplete or noisy annotations in real data, and (ii) it introduces greater diversity, particularly in terms of document layouts and structural variability.


Our synthetic generator is integrated throughout the entire training process, thanks to the strong resemblance between synthetic and real samples. This integration is designed to be non-blocking: data generation follows an asynchronous workflow, allowing training to proceed without delay. The training progressively shifts toward real data as it advances, with the proportion of real documents increasing over time.


The generator produces highly diverse and realistic newspaper pages, making it a valuable resource for training and evaluating document understanding systems. It introduces significant layout diversity, with a wide range of spatial configurations for sections, articles, and visual elements. The content and media are equally varied: articles are generated from French Wikipedia entries and enriched with images and advertisements sampled from a dedicated dataset of synthetic visuals. To better reflect real-world conditions, the generator also simulates common image degradations, such as blur, transparency, smudges, and stains, replicating artifacts introduced during printing and scanning. Finally, the rendering process is optimized to closely match the appearance of digitized newspapers, ensuring that the generated pages can be seamlessly used in training pipelines.





\

\section{The Finlam La Liberté Newspaper Dataset}
\label{sec:dataset}
We introduce the \textbf{Finlam La Liberté Newspaper Dataset}, a richly annotated, open-access\footnote{Available at \url{https://huggingface.co/datasets/Teklia/Newspapers-finlam-La-Liberte}} resource curated by the French National Library. It addresses the complex visual and semantic challenges found in 19th- and 20th-century newspapers, and is designed for training and evaluation end-to-end models in historical newspaper analysis.

\begin{figure}[ht]
    \centering
    \begin{subfigure}[b]{0.16\linewidth}
        \includegraphics[width=\linewidth]{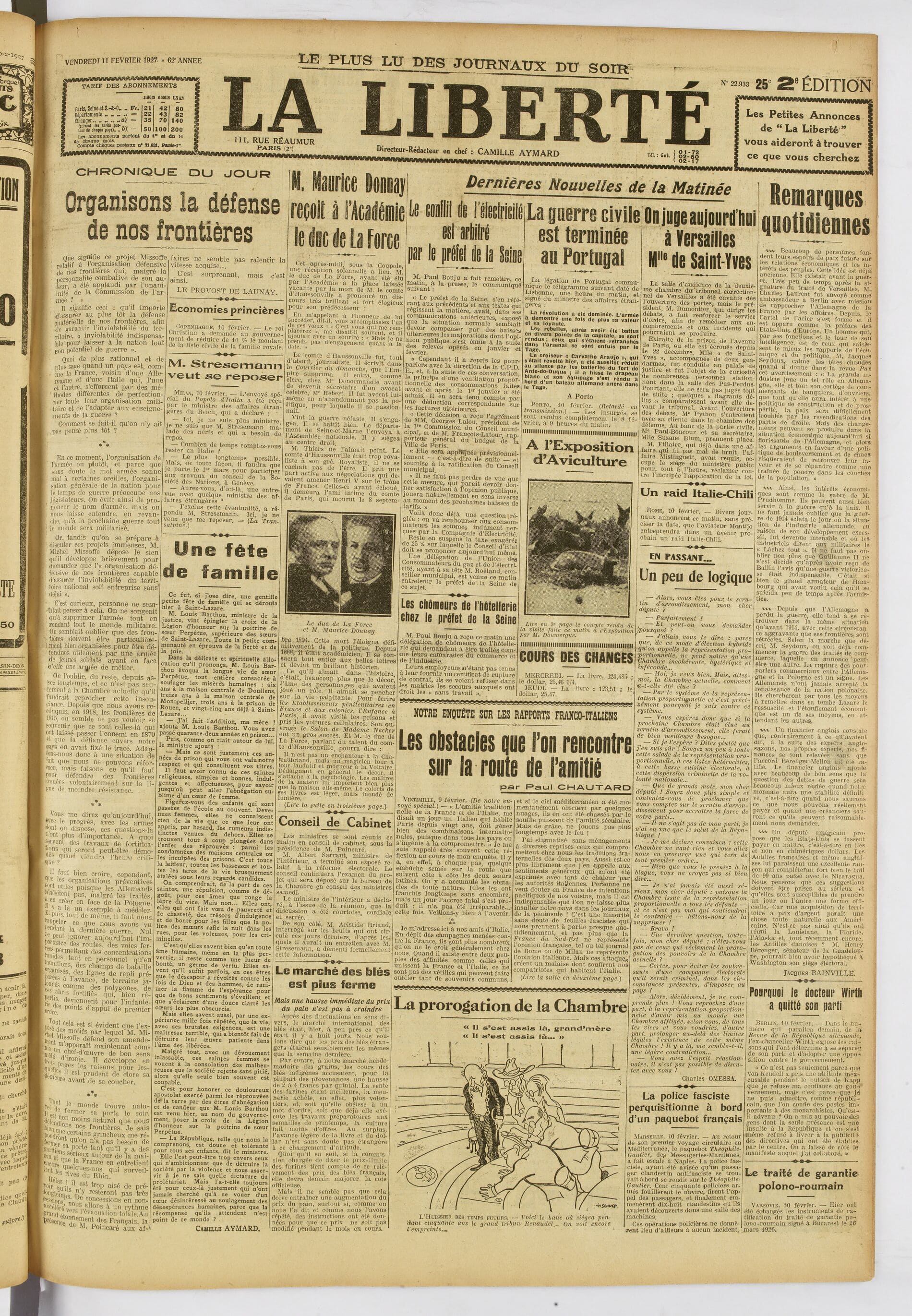}
    \end{subfigure}
    \hfill
    \begin{subfigure}[b]{0.16\linewidth}
        \includegraphics[width=\linewidth]{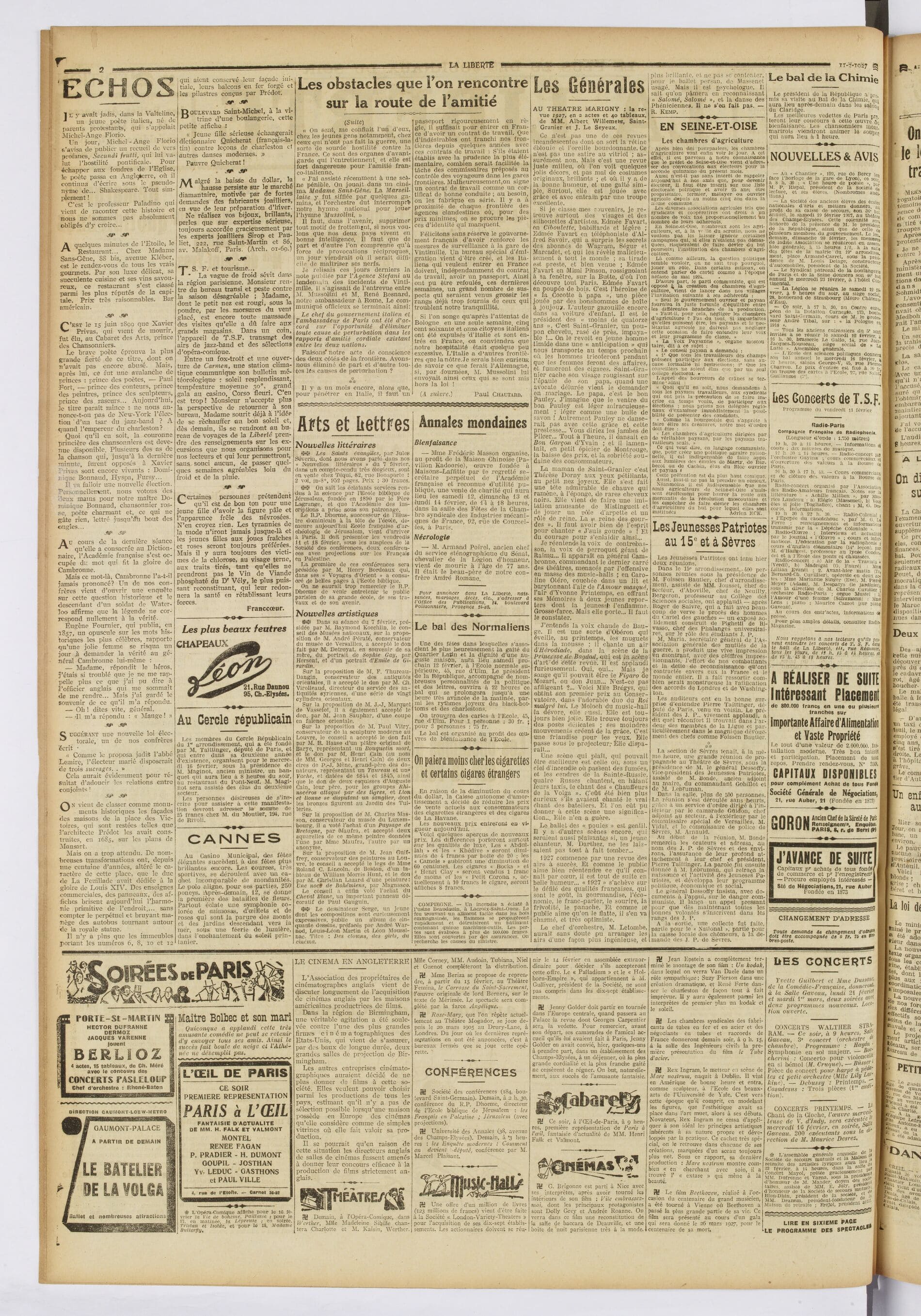}
    \end{subfigure}
    \hfill
    \begin{subfigure}[b]{0.16\linewidth}
        \includegraphics[width=\linewidth]{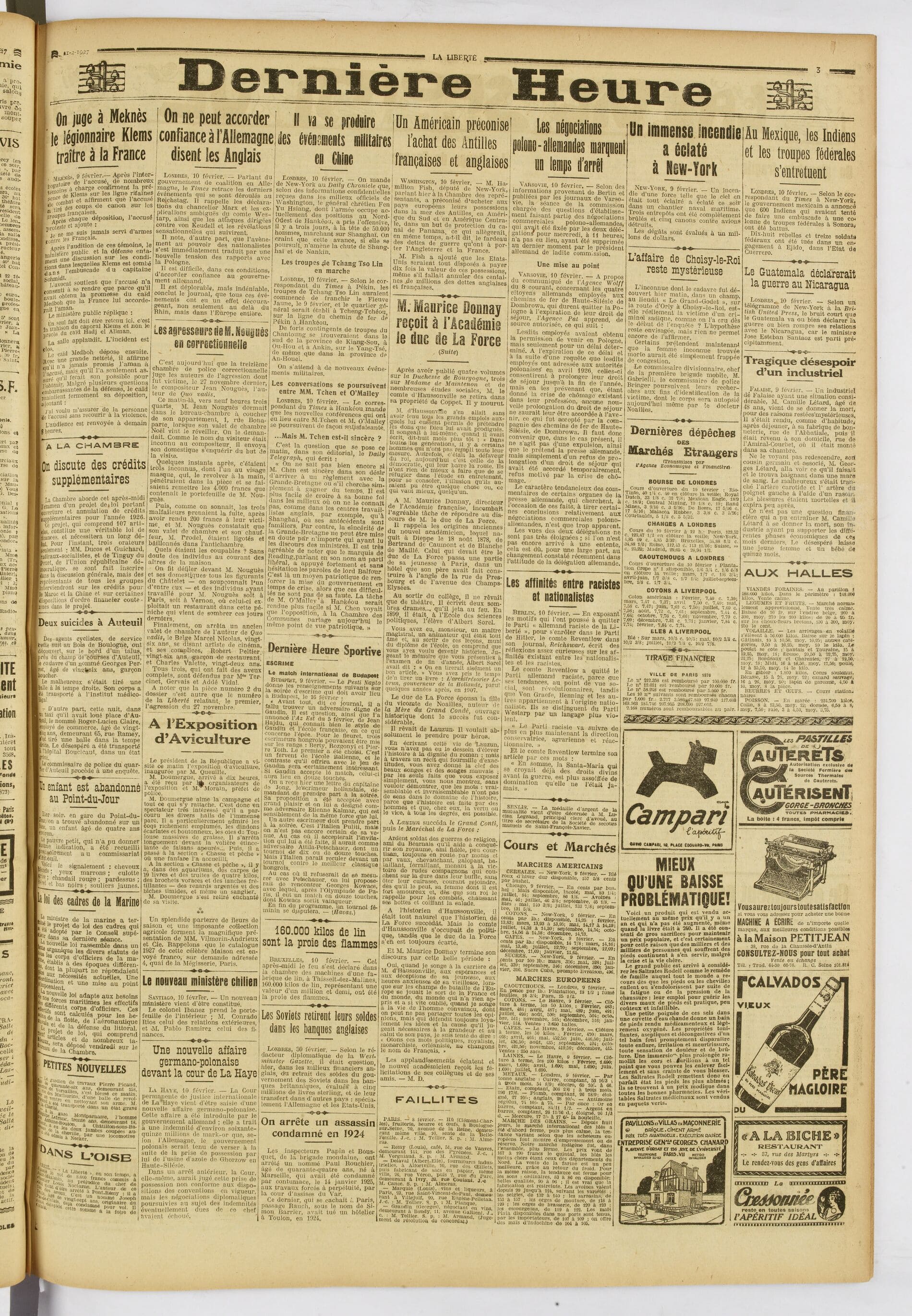}
    \end{subfigure}
    \hfill
    \begin{subfigure}[b]{0.16\linewidth}
        \includegraphics[width=\linewidth]{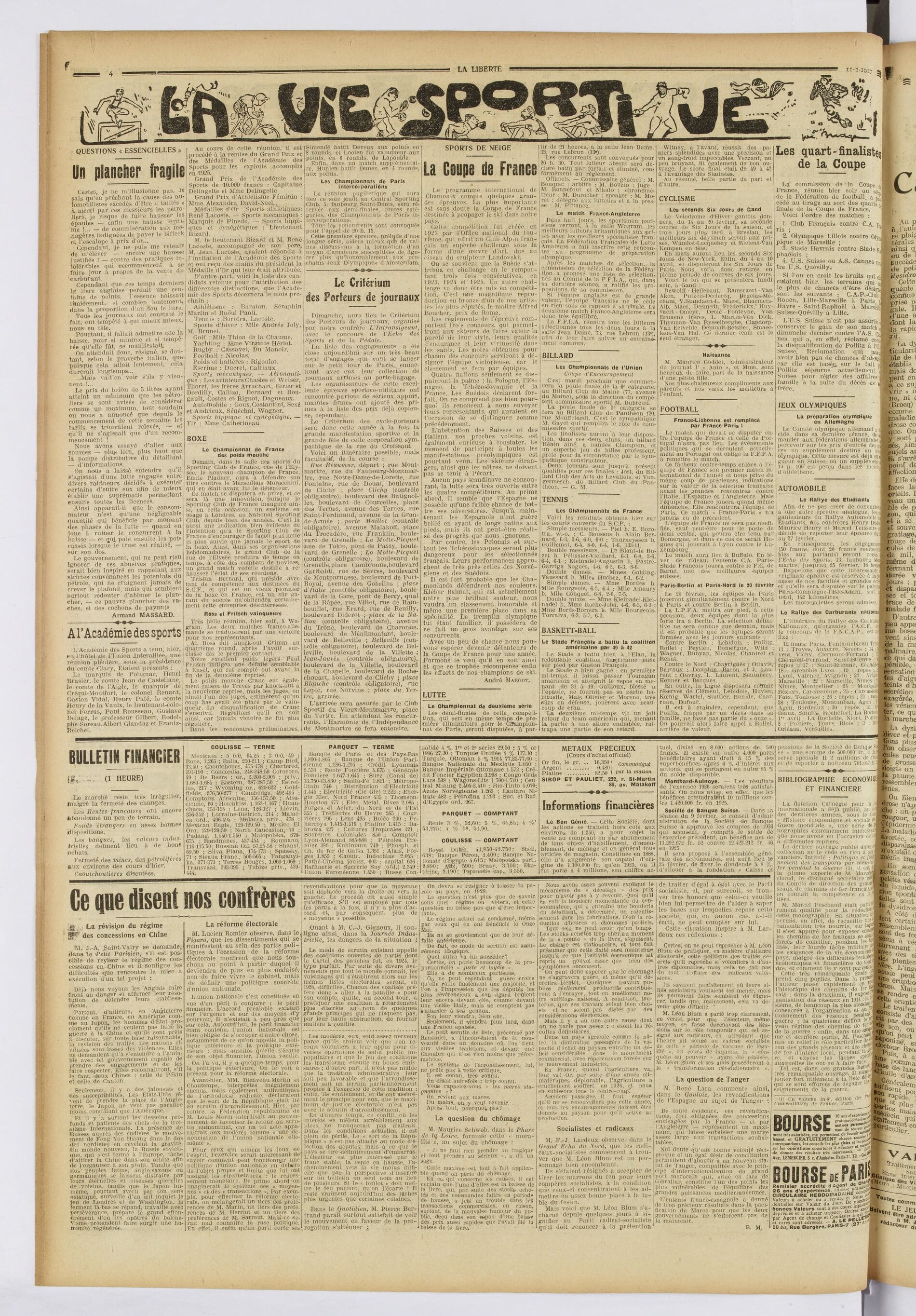}
    \end{subfigure}
    \hfill
    \begin{subfigure}[b]{0.16\linewidth}
        \includegraphics[width=\linewidth]{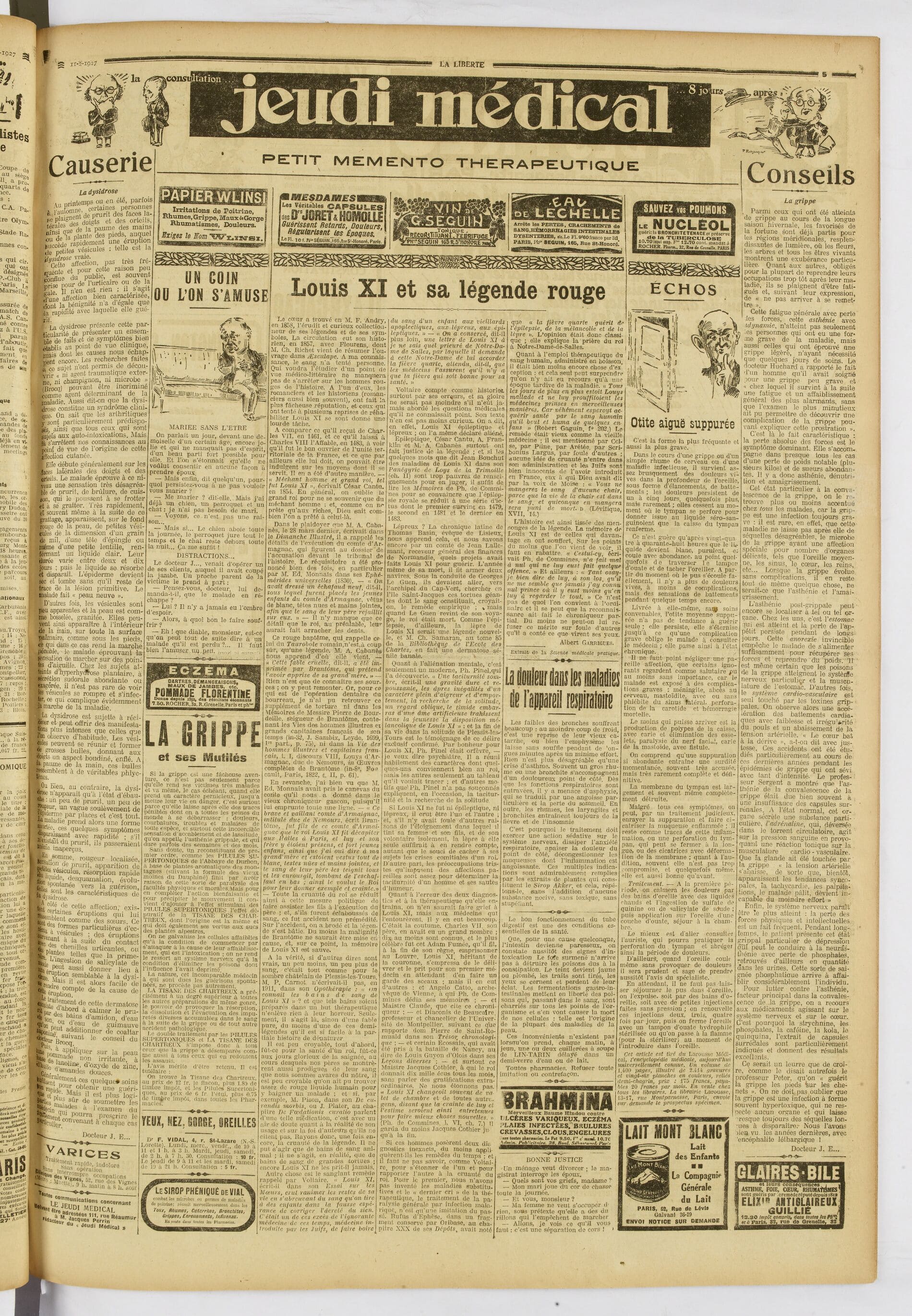}
    \end{subfigure}
    \hfill
    \begin{subfigure}[b]{0.16\linewidth}
        \includegraphics[width=\linewidth]{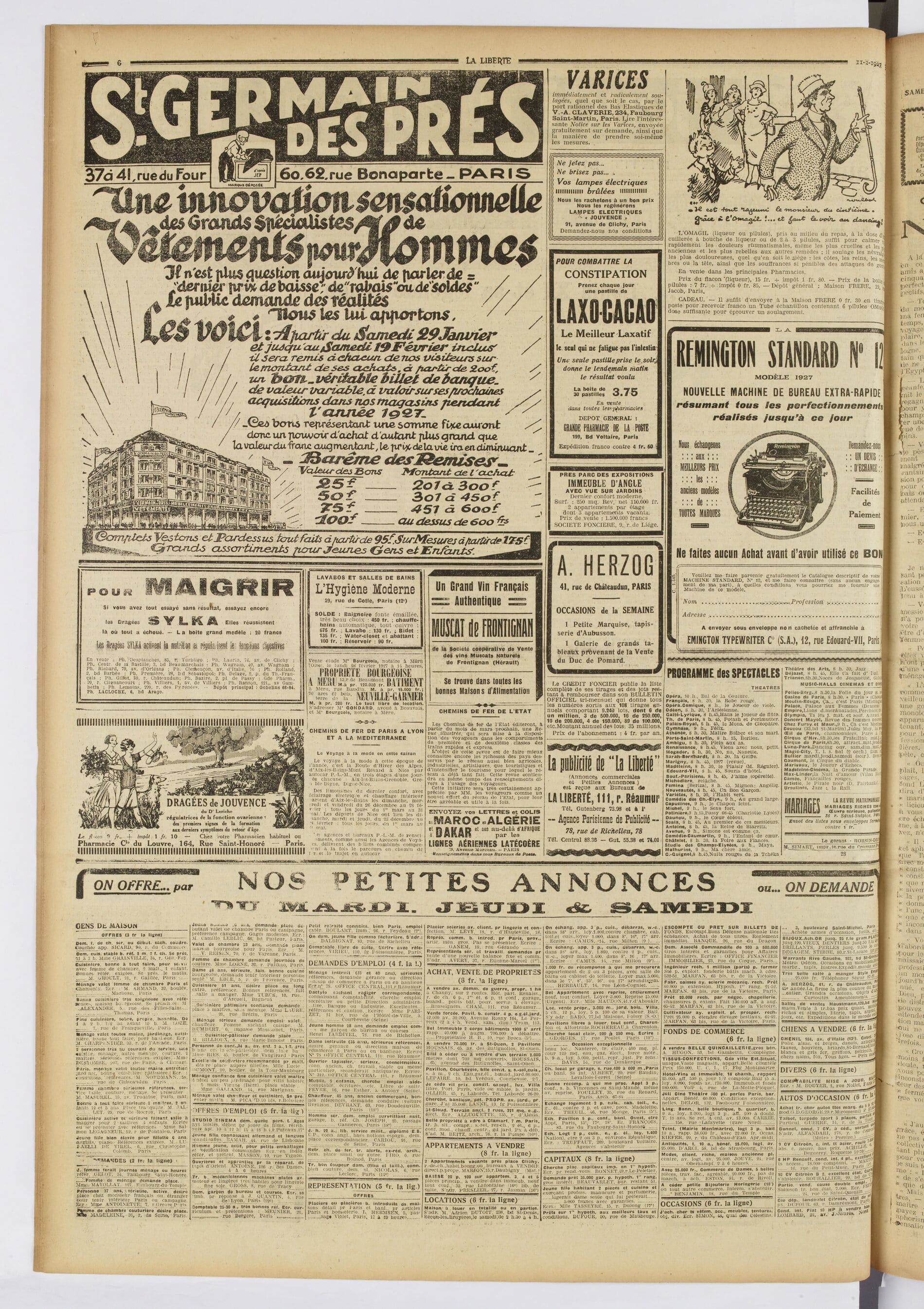}
    \end{subfigure}
    \caption{Examples of a complete issue of La Liberté dated February 11, 1927.}
    \label{fig:dataset_example}
\end{figure}

Finlam La Liberté includes \textbf{1500} complete issues from \textit{La Liberté}, a French-language newspaper. Each issue contains between \textbf{2 and 12} pages, with one high-resolution image per page, uniformly resized to a height of \textbf{2500 pixels} for consistency. Figure~\ref{fig:dataset_example} shows a full issue. 

Each page is segmented into \textit{semantic zones} and annotated with a rich set of \textit{multimodal features}, including \textbf{zone localization} via bounding boxes,  \textbf{OCR-extracted text content}, \textbf{classification} at multiple levels of granularity, and \textbf{article and section separation}.

The classification follows a hierarchical label set:
\begin{itemize}
    \item \textbf{Title Section classes} ($\mathcal{C}_{\texttt{TS}}$), such as \texttt{HEADER-TITLE} and \texttt{HEADER-TEXT} are used to characterize prominent layout components like newspaper headers.
    
    \item \textbf{Head-level classes} ($\mathcal{C}_{\texttt{head}}$) define the structure of article and section introductions, with labels including
         \texttt{SECTION-TITLE}, \texttt{SECTION-SUBTITLE},\\ \texttt{SECTION-ILLUSTRATEDTEXT},
         \texttt{TITLE}, \texttt{SUBTITLE}, \texttt{ILLUSTRATEDTEXT}

    \item \textbf{Body-level classes} ($\mathcal{C}_{\texttt{body}}$) capture the content within articles and sections, such as \texttt{TEXT}, \texttt{ILLUSTRATION}, \texttt{CAPTION}, \texttt{TABLE}, \texttt{AUTHOR}, and \texttt{INSIDEHEADING}

    \item \textbf{Article-level classes} ($\mathcal{C}_{\texttt{art}}$) describe the nature of each article, including \texttt{ARTICLE}, \texttt{ADVERTISEMENT}, and \texttt{FREEAD}. 
      \end{itemize}

The Finlam La Liberté dataset poses a substantial challenge due to the \textit{dense layouts}, \textit{diverse design templates}, and \textit{visual variability} across the issues of \emph{La Liberté}, including day-to-day layout fluctuations, evolving advertising formats, and typographic degradation typical of the printed press of that period. It offers a valuable benchmark for developing and evaluating multimodal systems capable of large-scale interpretation of complex historical print media.

However, some limitations remain: the OCR quality doesn't match current state-of-the-art models, and though segmentation and reading order are mostly reliable, specific pages contain errors. A gold standard dataset is in development to resolve this.

\subsection{Splits and training setup}
\label{sec:splits}

We use the official split released with the dataset: 1{,}350 issues (7{,}957 pages)
for training, 75 issues (446 pages) for validation, and 75 issues (433 pages)
for test\footnote{\url{https://huggingface.co/datasets/Teklia/Newspapers-finlam-La-Liberte}}.
Splits are defined at the issue level to avoid leakage between adjacent pages of the same day.

The three systems compared in Section~\ref{sec:evaluation} are trained on this split as follows.
\textbf{YOLO26} is trained on the training pages of Finlam La Liberté with the hyperparameters detailed in Section~\ref{sec:pipeline}.
\textbf{LayoutReader} is fine-tuned on the same training issues; it does not see the page images but only the normalized block coordinates
(rescaled to $[0, 1000]$), the predicted YOLO classes, and the vertical/horizontal separators detected by LSD.
\textbf{Tiramisu} is trained with a curriculum mixing synthetic and real
pages: the proportion of real data starts at $1\%$ and is increased by
$1\%$ per epoch, up to a cap of $\sim$$40\%$. This upper bound reflects the fact that real annotations are noisier and lexically less diverse than the synthetic pages, so a majority of clean and diverse synthetic supervision is retained throughout training. The synthetic data are generated on the fly by the generator described in Section~\ref{sec:tiramisu} and mixed asynchronously with the real batches, so that generation does not block training.

\section{Evaluation}
\label{sec:evaluation}
\subsection{Metrics}
All metrics are integrated into an evaluation framework that we developed, which is available here: \url{https://gitlab.teklia.com/adr/newspaper/evaluation}.

\subsubsection{Block detection and classification}
\label{metric:detection}
We rely on the \textbf{mAP} metric to evaluate the segmentation and classification quality of unit blocks. The Average Precision (AP) of each class is computed by finding the area under the Precision-Recall Curve, which plots precision against recall at various confidence thresholds. The class APs are then averaged to get the mean AP (mAP).

Two scores are computed:
\begin{itemize}
    \item \textbf{mAP@50} considers a detection correct if the Intersection over Union (IoU) is at least 0.50;
    \item \textbf{mAP@50-95} is the average of mAP calculated at IoU thresholds from 0.50 to 0.95 in steps of 0.05.
\end{itemize}

\subsubsection{Article separation quality}
\label{metric:sas}
The \textbf{AS} (Article Separation) metric evaluates the quality of the segmentation at the article level. The evaluation is based on the comparison between predicted and ground-truth masks, where each mask represents a single article as the union of its constituent blocks.

To align predictions with ground-truth instances, we use the Hungarian algorithm to find an optimal one-to-one matching that maximizes overlap. Matches are considered valid if the IoU between the predicted and ground-truth masks exceeds a given threshold (e.g., IoU $\geq 0.5$ for \texttt{AS@0.5}).

Once the matching is established, we compute the number of True Positives (TP), False Positives (FP), and False Negatives (FN) to derive standard performance metrics, Precision, Recall and \textbf{F1-score}.

In addition to these discrete metrics, we also report the mean IoU (\textbf{mIoU}) over the matched pairs identified by the Hungarian algorithm. This continuous measure captures the average quality of spatial alignment between matched predictions and ground-truth instances.




\subsubsection{Reading order with BLEU}

Following prior work~\cite{zhang2023}, we use the \textbf{BLEU} metric to assess the quality of predicted reading orders.

At the section level, we first align predicted and ground truth elements using the same matching strategy as for the \textbf{AS} metric (see Section~\ref{metric:sas}). \textbf{BLEU}~\cite{papineni2002bleu} is then computed over the indices of the matched elements, treating the order of elements as a sequence to be compared against the reference. This metric provides insight into the correctness of the predicted reading order. However, since unmatched elements are ignored, it should be interpreted jointly with \textbf{AS} to obtain a complete view of structure reconstruction quality.

At the block level, we flatten all blocks from the hierarchy into a single sequence and compute BLEU over the class labels, independent of spatial position. This approach avoids over-penalizing structural variations such as paragraph splitting or merging, which may occur without affecting the semantic flow much.

Finally, to evaluate the structural accuracy of the predictions at various hierarchical levels, we use a \textbf{Count Jaccard Index}, defined as:
\[
\texttt{Jaccard}(c_{\text{true}}, c_{\text{pred}}) = \frac{\min(c_{\text{true}}, c_{\text{pred}})}{\max(c_{\text{true}}, c_{\text{pred}})}
\]
where \( c_{\text{true}} \) and \( c_{\text{pred}} \) denote the number of ground truth and predicted elements, respectively. This index ranges from 0 (worst) to 1 (best), and offers a simple and effective measure of count alignment without requiring exact matching of positions. We compute this metric at the page level, considering the sequence of blocks (\textbf{BLEU$_{block}$}) and articles (\textbf{BLEU$_{article}$}).

\subsection{Results}

\subsubsection{Block detection and classification}

Table~\ref{tab:detection_results} compares the performance of all methods on the task of block detection and classification. The proposed bottom-up pipeline achieves the highest performance, significantly outperforming other methods, with a mAP@50 of 72.27\%. The Tiramisu approach comes second, reaching 39.14\% mAP@50, as its hierarchical processing passes can lead to cascading misses. For example, if an article or section is missed in the early passes, all of its constituent children blocks are also missed.

Finally, Arcanum comes last, which was expected, as it was not fine-tuned on this dataset. It should be noted that Arcanum tends to produce larger zones, while Tiramisu and the pipeline were both trained to create a new zone at each indented line. Moreover, because our classification schema is highly detailed, mapping Arcanum’s native classes to it inherently introduces classification mismatches that heavily penalize strict overlap metrics. From a qualitative perspective, Arcanum’s visual segmentations look acceptable, indicating that the low scores are driven by schema and granularity differences rather than failed detections.

\begin{table}[h]
  \centering
  \begin{tabular}{lcc}
    \toprule
    \textbf{Method} & \textbf{mAP@50 (\%)} & \textbf{mAP@50-95 (\%)}  \\ \midrule
    \textbf{Arcanum} & 11.12 & 6.48 \\ 
    \textbf{Tiramisu (top-down)} & 39.14 & 22.05 \\ 
    \textbf{Pipeline (bottom-up)}& \textbf{72.27} & \textbf{59.08}  \\ \bottomrule
  \end{tabular}
  \caption{Block detection and classification quality.}
  \label{tab:detection_results}
\end{table}

\subsubsection{Article detection quality}

Table~\ref{tab:sas_scores} evaluates article segmentation quality. Overall, the pipeline achieves the highest surface-based scores, with an F1 score of 80.39\% and a mIOU of 88.82\%, Tiramisu outperforms it on the count-based metrics (89.00\% Jaccard Article and 83.78\% Jaccard Section). However, Tiramisu outperforms the pipeline when it comes to the Jaccard Index. This suggests that the pipeline excels at surface-level precision, while Tiramisu’s top-down approach yields better overall entity counts.
Arcanum maintains a competitive mIOU (82.37\%) but falls behind in structural entity counting (50.22\% Jaccard Section) due to its zero-shot schema mismatches.

\begin{table}[h!]
  \centering
  \begin{tabular}{lcccc}
    \toprule
    \textbf{Method} & \textbf{F1 (\%)} & \textbf{mIOU (\%)} & \textbf{Jaccard$_{article}$ (\%)} & \textbf{Jaccard$_{section}$ (\%)} \\ \midrule
    \textbf{Arcanum} &  59.42 & 82.37 & 57.28 & 50.22 \\ 
    \textbf{Tiramisu (top-down)}  & 70.69 & 77.83 & \textbf{89.00} & \textbf{83.78} \\ 
    \textbf{Pipeline (bottom-up)} & \textbf{80.39} & \textbf{88.82} & 86.19 & 74.26 \\ \bottomrule
  \end{tabular}
  \caption{Article segmentation quality at AS@0.5.}
  \label{tab:sas_scores}
\end{table}

\subsubsection{Reading Order Detection Quality}

Table~\ref{tab:ro_scores} presents the performance of Tiramisu and the pipeline on the task of reading order detection. The proposed pipeline approach (bottom-up) achieves the highest block-level accuracy across the entire page, with a \textbf{BLEU$_{block}$} score of 87.20\%. This demonstrates the approach's precision in sequencing fine-grained layout elements globally. At the macro level, both the pipeline approach (97.08\%) and the Tiramisu top-down approach (97.43\%) achieve very high and comparable BLEU scores for articles, indicating that both models are highly effective at correctly ordering the sequence of articles on a page. Arcanum lags behind, especially at the block level, which makes sense based on the block segmentation results.

\begin{table}[h!]
  \centering
  \begin{tabular}{lcc}
    \toprule
    \textbf{Method} & \textbf{BLEU$_{block}$ (\%)} & \textbf{BLEU$_{article}$ (\%)} \\ \midrule
    \textbf{Arcanum} &  55.58 & 81.32 \\ 
    \textbf{Tiramisu (top-down)} & 73.31 & \textbf{97.43} \\ 
    \textbf{Pipeline (bottom-up)} & \textbf{87.20} & 97.08 \\ \bottomrule
  \end{tabular}
  \caption{Reading orders quality.}
  \label{tab:ro_scores}
\end{table}

\subsubsection{Speed}

In addition to structural quality, we evaluate the operational efficiency of the proposed approaches by measuring the average inference time per image. 

As shown in Table \ref{tab:speed}, the proposed bottom-up pipeline is faster than Tiramisu both on CPU (x34) and GPU (x1.75).

\begin{table}[]
    \centering
    \begin{tabular}{lcc}
    \toprule
         \textbf{Model} & \multicolumn{2}{c}{\textbf{Inference time per image (seconds)}}  \\
         & GPU (A100) & CPU (i5-12450H) \\
         \midrule
         \textbf{Tiramisu (top-down)} & 1.46 & 57.03 \\
         \textbf{Pipeline (bottom-up)} &  0.83 & 1.66 \\
         \bottomrule
    \end{tabular}
    \caption{Comparison of inference time for both approaches.}
    \label{tab:speed}
\end{table}

However, this increased speed and efficiency comes at the cost of greater system complexity. Tiramisu operates as a unified model, which makes it straightforward to deploy and update. In contrast, the bottom-up pipeline introduces a dependency chain consisting of three models.

\section{Conclusion}
\label{sec:conclusion}

In this paper, we described two approaches for Hierarchical Structure Understanding of Newspaper Images and release an open source dataset to train and evaluate them. 
First, we introduced a bottom-up pipeline, which chains separate models for layout analysis, reading-order prediction, and article reconstruction.
Second, we describe Tiramisu, an end-to-end model that jointly addresses these tasks within a single trainable architecture. It is the first end-to-end transformer-based model dedicated to the structured analysis of newspaper images. 

Our analysis reveals distinct trade-offs between the two approaches. Tiramisu offers an elegant end-to-end solution that handles all tasks—block detection and classification, article and section segmentation, OCR, and reading order—within a single model. However, it requires synthetic data for training and shows some detection misses despite producing high-quality zones when successful.
The bottom-up pipeline, combining YOLO with LSD for document layout analysis, LayoutReader for reading order, and post-processing for article segmentation, achieves better performance and faster inference. Its modular design allows flexibility, such as plugging in different models for each step, but introduces a dependency chain across four stages, making it more complex to deploy and maintain.

By the end of 2026, the French national Library will implement this bottom-up pipeline and test it on its extensive newspapers collection (over 8 million documents), with a view to deploying it on the largest possible scale.


%

\section*{Reproducibility}
The Finlam La Liberté dataset is publicly available at
\url{https://huggingface.co/datasets/Teklia/Newspapers-finlam-La-Liberte}.
The evaluation framework used to compute all the metrics reported in
Section~\ref{sec:evaluation} is released at
\url{https://gitlab.teklia.com/adr/newspaper/evaluation}. The Tiramisu
training code, including the synthetic newspaper generator, is available at
\url{https://git.litislab.fr/tiramisu/tiramisu-newspaper-articles-extractor}.

\section{Acknowledgment}
This work was performed using HPC computing resources from CRIANN (Normandy, France) and GENCI–IDRIS (Grant 2024-AD011015246). It has been funded by the French National Research Agency (ANR) under the agreement "Project-ANR-23-IAS1-0007".





 \bibliographystyle{splncs04}
 \bibliography{refs_short}






\end{document}